\documentclass{article} 
\usepackage{iclr2019_conference,times}

\usepackage{amsmath,amsfonts,bm}

\def\eqref#1{equation~\ref{#1}}

\def\1{\bm{1}}

\DeclareMathAlphabet{\mathsfit}{\encodingdefault}{\sfdefault}{m}{sl}
\SetMathAlphabet{\mathsfit}{bold}{\encodingdefault}{\sfdefault}{bx}{n}

\usepackage{hyperref}
\usepackage{url}
\usepackage{booktabs}
\usepackage{multirow}
\usepackage{array, caption, floatrow, makecell, booktabs}
\usepackage{wrapfig}
\usepackage{floatrow}
\usepackage{comment}

\newfloatcommand{capbtabbox}{table}[][\FBwidth]

\newcommand*{\rom}[1]{\uppercase\expandafter{\romannumeral #1\relax}}

\title{VisualBERT: A Simple and Performant\\ Baseline for Vision and Language}
{\centering
\author{Liunian Harold Li$^\dagger$, Mark Yatskar$^*$, Da Yin$^\circ$, Cho-Jui Hsieh$^\dagger$ \& Kai-Wei Chang$^\dagger$\\
$^\dagger$University of California, Los Angeles\\
$^*$Allen Institute for Artificial Intelligence\\
$^\circ$Peking University\\
\texttt{liunian.harold.li@cs.ucla.edu,  marky@allenai.org,}\\
\texttt{wade\_yin9712@pku.edu.cn, \{chohsieh, kwchang\}@cs.ucla.edu} \\
}
}

\usepackage{xspace,mfirstuc,tabulary}

\iclrfinalcopy 
\usepackage{todonotes}
\makeatletter
\newcommand*\iftodonotes{\if@todonotes@disabled\expandafter\@secondoftwo\else\expandafter\@firstoftwo\fi}  
\makeatother

\newcommand{\model}{VisualBERT\xspace}
\newcommand{\nlvr}{NLVR$^2$\xspace}
\newcommand{\bert}{BERT$_{\textsc{Base}}$\xspace}
\newcommand{\modelnp}{VisualBERT w/o COCO Pre-training\xspace}
\newcommand{\modelby}{VisualBERT w/o Early Fusion\xspace}
\newcommand{\modelp}{VisualBERT\xspace}

\begin{document}

\maketitle

\begin{abstract}
We propose VisualBERT, a simple and flexible framework for modeling a broad range of vision-and-language tasks. VisualBERT consists of a stack of Transformer layers that implicitly align elements of an input text and regions in an associated input image with self-attention. We further propose two visually-grounded language model objectives for pre-training VisualBERT on image caption data. Experiments on four vision-and-language tasks including VQA, VCR, \nlvr, and Flickr30K show that VisualBERT outperforms or rivals with state-of-the-art models while being significantly simpler.
Further analysis demonstrates that VisualBERT can ground elements of language to image regions without any explicit supervision and is even sensitive to syntactic relationships, tracking, for example, associations between verbs and image regions corresponding to their arguments.
\end{abstract}

\section{Introduction}

Tasks combining vision and natural language serve as a rich test-bed for evaluating the reasoning capabilities of visually informed systems.
Beyond simply recognizing what objects are present~\citep{russakovsky2015imagenet,lin2014microsoft}, vision-and-language tasks, such as captioning~\citep{chen2015microsoft}, visual question answering~\citep{antol2015vqa}, and visual reasoning~\citep{suhr2018corpus,zellers2019recognition}, challenge systems to understand a wide range of \textit{detailed semantics} of an image, including objects, attributes, parts, spatial relationships, actions and intentions, and how all of these concepts are referred to and grounded in natural language. 

In this paper, we propose \model, a simple and flexible model designed for capturing rich semantics in the image and associated text. \model integrates BERT~\citep{devlin2018bert}, a recent Transformer-based model~\citep{vaswani2017attention} for natural language processing, and pre-trained object proposals systems such as Faster-RCNN~\citep{ren2015faster} and it can be applied to a variety of vision-and-language tasks. In particular, image features extracted from object proposals are treated as unordered input tokens and fed into \model along with text. The text and image inputs are jointly processed  by multiple Transformer layers in \model (See Figure~\ref{fig:model}). The rich interaction among words and object proposals allows the model to capture the intricate associations between text and image.

Similar to BERT, pre-training \model on external resource can benefit downstream applications. In order to learn associations between images and text, we consider pre-training \model on image caption data, where \textit{detailed semantics} of an image are expressed in natural language. 
We propose two \textit{visually-grounded} language model objectives for pre-training: (1) part of the text is masked and the model learns to predict the masked words based on the remaining text and visual context; (2) the model is trained to determine whether the provided text matches the image. We show that such pre-training on image caption data is important for \model to learn transferable text and visual representations. 

\begin{figure*} [t]
\centering
\includegraphics[width=1.0\textwidth]{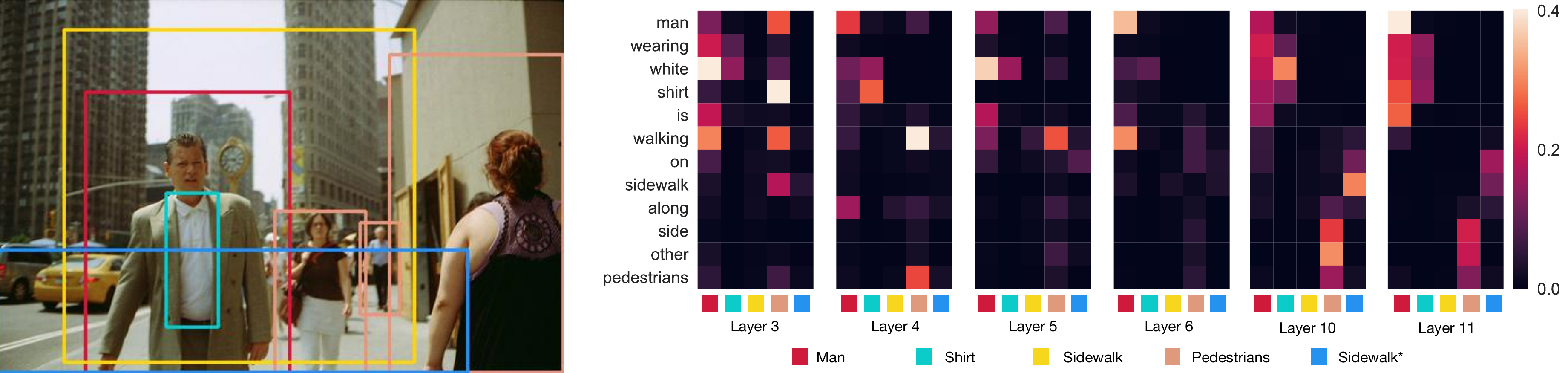}
\caption{Attention weights of some selected heads in \model. In high layers (e.g., the 10-th and 11-th layer), \model is capable of implicitly grounding visual concepts (e.g., ``other pedestrians'' and ``man wearing white shirt'').  
The model also captures certain syntactic dependency relations (e.g., ``walking'' is aligned to the \textit{man} region in the 6-th layer). The model also refines its understanding over the layers, incorrectly aligning ``man'' and ``shirt'' in the 3-rd layer but correcting them in higher layers.  (See more details in \S \ref{sec:casestudy}.) }
\label{fig:leading_example}
\end{figure*}

We conduct comprehensive experiments on four vision-and-language tasks: (1) visual question answering (VQA 2.0, \citet{balanced_vqa_v2}), (2) visual commonsense reasoning (VCR, \citet{zellers2019recognition}), (3) natural language for visual reasoning (NLVR$^2$, \citet{suhr2018corpus}), and (4) region-to-phrase grounding (Flickr30K, \citet{plummer2015flickr30k}).
Results demonstrate that by pre-training \model on the COCO image caption dataset \citep{chen2015microsoft}, \model outperforms or rivals with the state-of-the-art models. We further provide detailed ablation study to justify our design choices. Further quantitative and qualitative analysis reveals how \model allocates attention weights to align words and image regions internally. We demonstrate that through pre-training, \model learns to ground entities and encode certain dependency relationships between words and image regions, which attributes to improving the model's understanding on the detailed semantics of an image (see an example in Figure \ref{fig:leading_example}).

\section{Related Work}
\label{sec:related}
There is a long research history of bridging vision and language. Various tasks such as visual question answering~\citep{antol2015vqa,balanced_vqa_v2}, textual grounding~\citep{kazemzadeh2014referitgame,plummer2015flickr30k}, and visual reasoning~\citep{suhr2018corpus,zellers2019recognition} have been proposed and various models~\citep{yang2016stacked,anderson2018bottom,jiang2018pythia} have been developed to solve them. These approaches often consist of a text encoder, an image feature extractor, a multi-modal fusion module (typically with attention), and an answer classifier. Most models are designed for specific tasks, while \model is general and can be easily adapted to new tasks or incorporated into other task-specific models.

Understanding \textit{detailed semantics} depicted in an image is critical for visual understanding~\citep{johnson2015image} and prior studies show that modeling such semantics can benefit visual-an-language models. 
For instance, attribute annotations in Visual Genome~\citep{krishna2017visual} are used to enhance the object detector in VQA systems~\citep{anderson2018bottom}. \citet{santoro2017simple}, \citet{norcliffe2018learning}, and \citet{cadene2019murel} explore using an attention module to implicitly model the relations between objects in the image. \citet{Li2019RelationawareGA} take a further step and explicitly build a graph to encode object relations. In \model, the self-attention mechanism allows the model to capture the implicit relations between objects. Furthermore, we argue that pre-training on image caption data is an effective way to teach the model how to capture such relations.

Our work is inspired by BERT~\citep{devlin2018bert}, a Transformer-based representation model for natural language.
It falls into a line of works~\citep{peters2018deep,radford2018improving,radford2019language} that learn a universal language encoder by pre-training with language modeling objective (i.e., predicting words that are masked out from the input based on the remaining context).
Two concurrent studies resemble this paper. VideoBERT~\citep{sun2019videobert} transforms a video into spoken words paired with a series of images and applies a Transformer to learn joint representations.
Their model architecture is similar to ours.
However, VideoBERT is evaluated on captioning for cooking videos, while we conduct comprehensive analysis on a variety of vision-and-language tasks. 
Concurrently with our work, ViLBERT~\citep{lu2019vilbert} proposes to learn joint representation of images and text using a BERT-like architecture but has separate Transformers for vision and language that can only attend to each-other (resulting in twice the parameters). They use a slightly different pre-training process on Conceptual Captions~\citep{sharma2018conceptual} and conduct evaluation on four datasets, two of which are also considered in our work. Our results are consistent with theirs (our model outperforms on one out of the two intersecting tasks), but the methods are not wholly comparable because different visual representation and pre-training resource are used.

\section{A Joint Representation Model for Vision and Language}\label{approach}

\begin{figure*} [t]
\centering
\includegraphics[width=\textwidth]{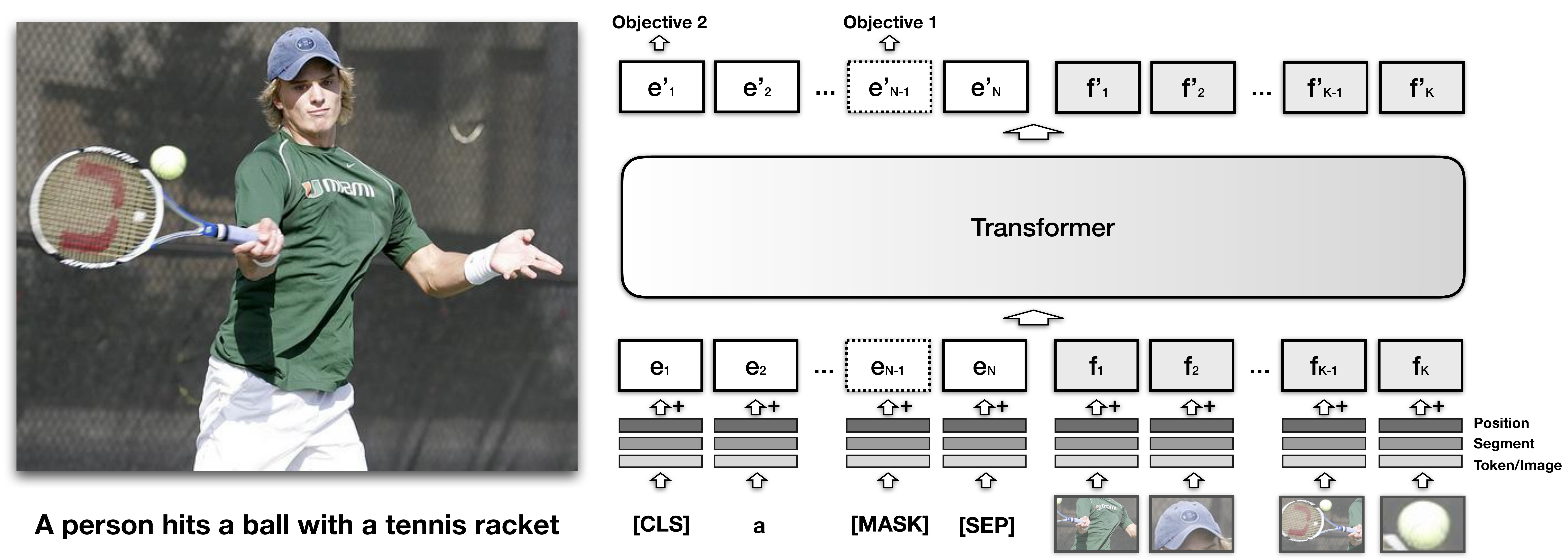}
\caption{The architecture of VisualBERT. Image regions and language are combined with a Transformer to allow the self-attention to discover implicit alignments between language and vision. It is pre-trained with a masked language modeling (Objective 1), and sentence-image prediction task (Objective 2), on caption data and then fine-tuned for different tasks. See \S \ref{training_process} for more details.
}
\label{fig:model}
\end{figure*}
In this section we introduce \model, a model for learning joint contextualized representations of vision and language.
First we give background on BERT (\S \ref{subsec:background}), then summarize the adaptations we made to allow processing images and text jointly (\S \ref{subsec:model}), as seen in Figure~\ref{fig:model}, and finally explain our training procedure (\S ~\ref{subsec:training}).

\subsection{Background}
\label{subsec:background}
BERT~\citep{devlin2018bert} is a Transformer~\citep{vaswani2017attention} with subwords~\citep{wu2016google} as input and trained using language modeling objectives.
All of the subwords in an input sentence are mapped to a set of embeddings, $E$. Each embedding $e \in E$ is computed as the sum of 1) a token embedding $e_t$, specific to the subword, 2) a segment embedding $e_s$, indicating which part of text the token comes from (e.g., the hypothesis from an entailment pair) and 3) a position embedding $e_p$, indicating the position of the token in the sentence.
The input embeddings $E$ are then passed through a multi-layer Transformer that builds up a contextualized representation of the subwords.

BERT is commonly trained with two steps: pre-training and fine-tuning.
Pre-training is done using a combination of two language modeling objectives: (1) masked language modeling, where some parts of the input tokens are randomly replaced with a special token (i.e., [MASK]), and the model needs to predict the identity of those tokens and (2) next sentence prediction, where the model is given a sentence pair and trained to classify whether they are two consecutive sentences from a document.
Finally, to apply BERT to a particular task, a task-specific input, output layer, and objective are introduced, and the model is fine-tuned on the task data from pre-trained parameters.

\subsection{VisualBERT}
\label{subsec:model}
\label{model}

The core of our idea is to reuse the self-attention mechanism within the Transformer to implicitly align elements of the input text and regions in the input image.
In addition to all the components of BERT, we introduce a set of visual embeddings, $F$, to model an image.
Each $f \in F$ corresponds to a bounding region in the image, derived from an object detector.

Each embedding in $F$ is computed by summing three embeddings: (1) $f_o$, a visual feature representation of the bounding region of $f$, computed by a convolutional neural network, (2) $f_s$, a segment embedding indicating it is an image embedding as opposed to a text embedding, and (3) $f_p$, a position embedding, which is used when alignments between words and bounding regions are provided as part of the input, and set to the sum of the position embeddings corresponding to the aligned words (see VCR in \S \ref{experiment}). 
The visual embeddings are then passed to the multi-layer Transformer along with the original set of text embeddings, allowing the model to implicitly discover useful alignments between both sets of inputs, and build up a new joint representation.\footnote{If text and visual input embeddings are of different dimension, we project the visual embeddings into a space of the same dimension as the text embeddings.}

\subsection{Training \model}
\label{training_process}
\label{subsec:training}

We would like to adopt a similar training procedure as BERT but \model must learn to accommodate both language and visual input.
Therefore we reach to a resource of paired data: COCO~\citep{chen2015microsoft} that contains images each paired with 5 independent captions. Our training procedure contains three phases:

\vspace{-5pt}
\paragraph{Task-Agnostic Pre-Training}
Here we train VisualBERT on COCO using two \textit{visually-grounded} language model objectives. 
(1) Masked language modeling with the image. Some elements of text input are masked and must be predicted but vectors corresponding to image regions are not masked. 
(2) Sentence-image prediction. For COCO, where there are multiple captions corresponding to one image, we provide a text segment consisting of two captions. One of the caption is describing the image, while the other has a 50\% chance to be another corresponding caption and a 50\% chance to be a randomly drawn caption. The model is trained to distinguish these two situations.

\vspace{-5pt}
\paragraph{Task-Specific Pre-Training} Before fine-tuning \model to a downstream task, we find it beneficial to train the model using the data of the task with the masked language modeling with the image objective. This step allows the model to adapt 
to the new target domain.

\vspace{-5pt}
\paragraph{Fine-Tuning}
This step mirrors BERT fine-tuning, where a task-specific input, output, and objective are introduced, and the Transformer is trained to maximize performance on the task.
\section{Experiment}
\label{experiment}
We evaluate \model on four different types of vision-and-language applications:
(1) Visual Question Answering (VQA 2.0)~\citep{balanced_vqa_v2}, (2) Visual Commonsense Reasoning (VCR)~\citep{zellers2019recognition},
(3) Natural Language for Visual Reasoning (\nlvr)~\citep{suhr2018corpus}, and (4) Region-to-Phrase Grounding (Flickr30K)~\citep{plummer2015flickr30k}, each described in more details in the following sections and the appendix.
For all tasks, we use the Karpathy train split~\citep{karpathy2015deep} of COCO for task-agnostic pre-training, which has around 100k images with 5 captions each. The Transformer encoder in all models has the same configuration as \bert: 12 layers, a hidden size of 768, and 12 self-attention heads. The parameters are initialized from the pre-trained \bert parameters released by \citet{devlin2018bert}.

For the image representations, each dataset we study has a different standard object detector to generate region proposals and region features. To compare with them, we follow their settings, and as a result, different image features are used for different tasks (see details in the subsections).~\footnote{Ideally, we can use the best available detector and visual representation for all tasks, but we would like to compare methods on similar footing.} For consistency, during task-agnostic pre-training on COCO, we use the same image features as in the end tasks.
For each dataset, we evaluate three variants of our model: 

\textbf{\modelp}: The full model with parameter initialization from BERT that undergoes pre-training on COCO, pre-training on the task data, and fine-tuning for the task. 

\textbf{\modelby}: \model but where image representations are not combined with the text in the initial Transformer layer but instead at the very end with a new Transformer layer. 
This allows us to test whether interaction between language and vision throughout the whole Transformer stack is important to performance.

\textbf{\modelnp}: \model but where we skip task-agnostic pre-training on COCO captions. This allows us to validate the importance of this step.

Following \citet{devlin2018bert}, we optimize all models using SGD with Adam~\citep{kingma2014adam}. We set the warm-up step number to be 10\% of the total training step count unless specified otherwise. Batch sizes are chosen to meet hardware constraints and text sequences whose lengths are longer than 128 are capped. Experiments are conducted on Tesla V100s and GTX 1080Tis, and all experiments can be replicated on at most 4 Tesla V100s each with 16GBs of GPU memory. Pre-training on COCO generally takes less than a day on 4 cards while task-specific pre-training and fine-tuning usually takes less. 
Other task-specific training details are in the corresponding sections.

\subsection{VQA}
Given an image and a question, the task is to correctly answer the question. 
We use the VQA 2.0~\citep{balanced_vqa_v2}, consisting of over 1 million questions about images from COCO.
We train the model to predict the 3,129 most frequent answers and use image features from a ResNeXt-based Faster RCNN pre-trained on Visual Genome~\citep{jiang2018pythia}.
More details are in Appendix \ref{appendix:vqa}.

We report the results in Table \ref{vqa}, including baselines using the same visual features and number of bounding region proposals as our methods (first section), our models (second section), and other incomparable methods (third section) that use external question-answer pairs from Visual Genome (+VG) , multiple detectors~\citep{yu2019multimodal} (+Multiple Detectors) and ensembles of their models.
In comparable settings, our method is significantly simpler and outperforms existing work.

\begin{table}[h]
\small
\caption{Model performance on VQA. \model outperforms Pythia v0.1 and v0.3, which are tested under a comparable setting.}
\label{vqa}
\begin{center}
\begin{tabular}{l|cccc}
\toprule

Model & Test-Dev & Test-Std \\ 
\midrule
Pythia v0.1~\citep{jiang2018pythia} & 68.49 & - \\
Pythia v0.3~\citep{singh2019towards} & 68.71 & - \\
\midrule
\modelby & 68.18 & - \\
\modelnp & 70.18 & - \\
\modelp & 70.80 & 71.00\\
\midrule
\midrule
Pythia v0.1 + VG + Other Data Augmentation~\citep{jiang2018pythia} & 70.01 & 70.24\\

MCAN + VG ~\citep{yu2019deep} & 70.63 & 70.90 \\
MCAN + VG + Multiple Detectors~\citep{yu2019deep} & 72.55 & - \\
MCAN + VG + Multiple Detectors + BERT~\citep{yu2019deep} & 72.80 & - \\
MCAN + VG + Multiple Detectors + BERT + Ensemble~\citep{yu2019deep} & 75.00 & 75.23 \\
\bottomrule
\end{tabular}
\end{center}
\end{table}

\subsection{VCR}
\label{vcr_experiment}
VCR consists of 290k questions derived from 110k movie scenes, where the questions focus on visual commonsense. The task is decomposed into two multi-choice sub-tasks wherein we train individual models: question answering (Q $\rightarrow$ A) and answer justification (QA $\rightarrow$ R). 
Image features are obtained from a ResNet50~\citep{he2016deep} and ``gold'' detection bounding boxes and segmentations provided in the dataset are used\footnote{In the fine-tuning stage, for \model (with/without Early Fusion), ResNet50 is fine-tuned along with the model as we find it beneficial. For reference, \modelp with a fixed ResNet50 gets 51.4 on the dev set for Q $\rightarrow$ AR. The ResNet50 of \modelnp is not fine-tuned with the model such that we could compare it with R2C fairly.}. 
The dataset also provides alignments between words and bounding regions that are referenced to in the text, which we utilize by using the same position embeddings for matched words and regions. More details are in Appendix \ref{appendix:vcr}.

Results on VCR are presented in Table \ref{vcr}.
We compare our methods against the model released with the dataset which builds on BERT (R2C) and list the top performing single model on the leaderboard (B2T2). Our ablated \modelnp enjoys the same resource as R2C, and despite being significantly simpler, outperforms it by a large margin. The full model further improves the results. Despite substantial domain difference between COCO and VCR, with VCR covering scenes from movies, pre-training on COCO still helps significantly. 

\begin{table}[h]
\small
\caption{Model performance on VCR. \modelnp outperforms R2C, which enjoys the same resource while \model further improves the results.}
\label{vcr}
\begin{center}
\begin{tabular}{l|llllll}
\toprule

\multirow{2}{*}{Model} & \multicolumn{2}{c}{Q $\rightarrow$ A} & \multicolumn{2}{c}{QA $\rightarrow$ R} & \multicolumn{2}{c}{Q $\rightarrow$ AR} \\ 
  & Dev & Test & Dev & Test & Dev & Test  \\ 
\midrule

R2C \citep{zellers2019recognition} & 63.8 & 65.1 & 67.2 & 67.3 & 43.1 & 44.0 \\
B2T2 (Leaderboard; Unpublished) & -  & 72.6 & - &75.7 &- & 55.0 \\
\midrule

\modelby & 70.1 & - & 71.9 & - & 50.6 & - \\

\modelnp & 67.9 & - & 69.5 & - & 47.9 & - \\ 

\modelp & 70.8 & 71.6  & 73.2 & 73.2 & 52.2 & 52.4 \\

\bottomrule
\end{tabular}
\end{center}
\end{table}

\subsection{\nlvr}

\nlvr is a dataset for joint reasoning about natural language and images, with a focus on semantic diversity, compositionality, and visual reasoning challenges.
The task is to determine whether a natural language caption is true about a pair of images.
The dataset consists of over 100k examples of English sentences paired with web images.
We modify the segment embedding mechanism in \model and assign features from different images with different segment embeddings. We use an off-the-shelf detector from Detectron~\citep{Detectron2018} to provide image features and use 144 proposals per image.\footnote{We conducted a preliminary experiment on the effect of the number of object proposals we keep per image. We tested models with 9, 18, 36, 72, and 144 proposals, which achieve an accuracy of 64.8, 65.5, 66.7, 67.1, and 67.4 respectively on the development set.} More details are in Appendix \ref{appendix:nlvr}.

Results are in Table \ref{nlvr}. \modelby and \modelnp surpass the previous best model MaxEnt by a large margin while \modelp widens the gap.

\begin{table}[h]
\small
\begin{tabular}{l|llllll}
    \toprule
    Model  & Dev & Test-P & Test-U & Test-U (Cons)\\ 
    \midrule 

    MaxEnt \citep{suhr2018corpus} & 54.1 & 54.8 & 53.5 & 12.0 \\
    
    \midrule
    \modelby & 64.6 & - & - & - \\
    
    \modelnp & 63.5  & - & - & - \\

    \modelp & 67.4 & 67.0 & 67.3 & 26.9\\
    
    \bottomrule
    \end{tabular}
\caption{Comparison with the state-of-the-art model on \nlvr. The two ablation models significantly outperform MaxEnt while the full model widens the gap. } \label{nlvr}
\end{table}
\subsection{Flickr30K Entities}

Flickr30K Entities dataset tests the ability of systems to ground phrases in captions to bounding regions in the image. The task is, given spans from a sentence, selecting the bounding regions they correspond to. The dataset consists of 30k images and nearly 250k annotations. We adapt the setting of BAN~\citep{kim2018bilinear}, where image features from a Faster R-CNN pre-trained on Visual Genome are used.
For task specific fine-tuning, we introduce an additional self-attention block and use the average attention weights from each head to predict the alignment between boxes and phrases. For a phrase to be grounded, we take whichever box receives the most attention from the last sub-word of the phrase as the model prediction. More details are in Appendix \ref{appendix:flickr}.

Results are listed in Table \ref{table:flickr}.
\model outperforms the current state-of-the-art model BAN.
In this setting, we do not observe a significant difference between the ablation model without early fusion and our full model, arguing that perhaps a shallower architecture is sufficient for this task.

\begin{table}[h]
\caption{Comparison with the state-of-the-art model on the Flickr30K. \model holds a clear advantage over BAN.}
\small
\label{sample-table}
\begin{center}
\begin{tabular}{@{}l@{ }|@{ }ccccccccccc@{ }c@{}}
\toprule
\multirow{2}{*}{Model} & \multicolumn{2}{c}{R@1} & \multicolumn{2}{c}{R@5} & \multicolumn{2}{c}{R@10} & \multicolumn{2}{c}{Upper Bound} \\ 

 & Dev & Test & Dev & Test & Dev & Test & Dev & Test \\ 

\midrule

BAN~\citep{kim2018bilinear} & - & 69.69  & - & 84.22 & - & 86.35 & 86.97 & 87.45 \\

\midrule

\modelby & 70.33 & - & 84.53 & - &  86.39 & - & \multirow{3}{*}{86.97} & \multirow{3}{*}{87.45}\\

\modelnp & 68.07 & - & 83.98 & - & 86.24 & -  \\
\modelp & 70.40 & 71.33 & 84.49 & 84.98 & 86.31 & 86.51  \\

\bottomrule
\end{tabular}
\end{center}
\label{table:flickr}
\end{table}

\section{Analysis}
In this section we conduct extensive analysis on what parts of our approach are important to \model's strong performance (\S~\ref{abalation_study}). 
Then we use Flickr30K as a diagnostic dataset to understand whether \model's pre-training phase actually allows the model to learn implicit alignments between bounding regions and text phrases. 
We show that many attention heads within \model accurately track grounding information and that some are even sensitive to syntax, attending from verbs to the bounding regions corresponding to their arguments within a sentence (\S~\ref{sec:analysis}).
Finally, we show qualitative examples of how \model resolves ambiguous groundings through multiple layers of the Transformer (\S~\ref{sec:casestudy}).

\subsection{Ablation Study}
\label{abalation_study}
\begin{figure}[t]
\begin{floatrow}
\capbtabbox{
\resizebox{0.95\linewidth}{!}{
    \small
    \centering
    \begin{tabular}{@{}l@{ }l@{ }|@{ }llll@{}}
    \toprule
    & Model  & Dev \\
    \midrule
    & \model & 66.7 \\
    \midrule

    \multirow{2}{*}{C1} 
    &\modelp w/o Grounded Pre-training & 63.9 \\
    &\modelnp & 62.9 \\
    \midrule 
    
    \multirow{1}{*}{C2}
    & \modelby  & 61.4 \\
    \midrule
    
    \multirow{1}{*}{C3} &\modelp w/o BERT Initialization & 64.7 \\
    \midrule 
    \multirow{1}{*}{C4} &\modelp w/o Objective 2 & 64.9  \\
    \bottomrule
    \end{tabular}
    }
}{
    \caption{Performance of the ablation models on \nlvr. Results confirm that task-agnostic pre-training (C1) and early fusion of vision and language (C2) are essential for \model.} 
    \label{nlvr_ablation}
}
\ffigbox{
    \centering
  \includegraphics[width=\linewidth]{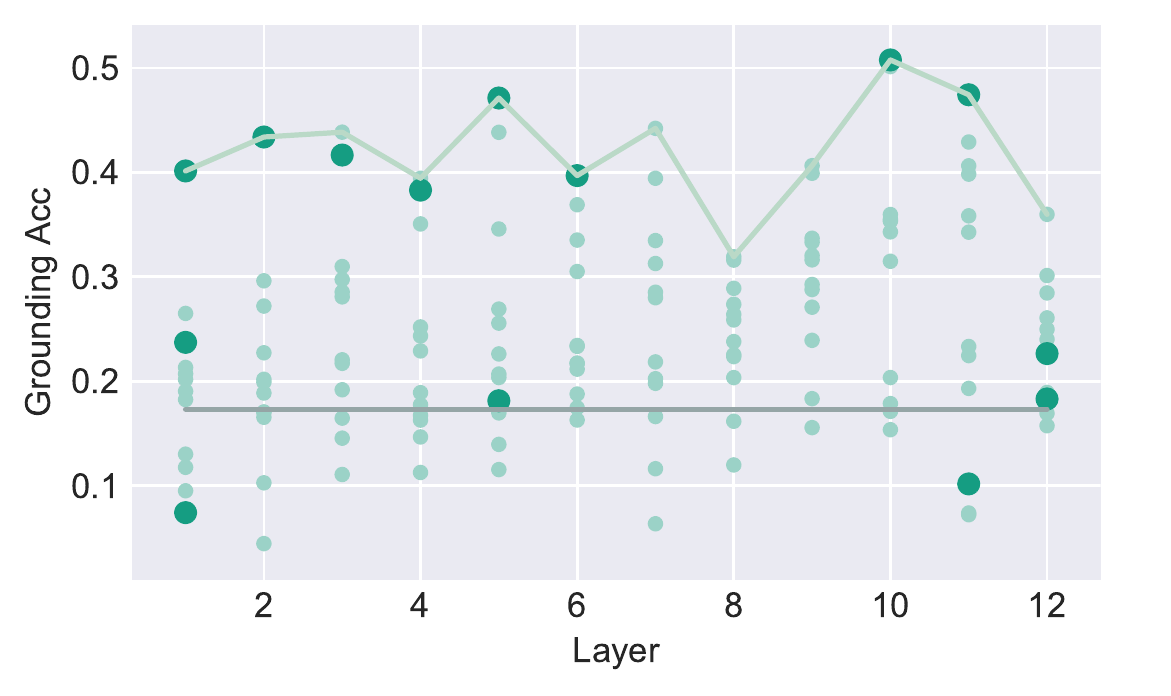}
  \vspace{-20pt}
  }{
    \caption{Entity grounding accuracy of the attention heads of \model. The rule-based baseline is dawn as the grey line. We find that certain heads achieves high accuracy while the accuracy peaks at higher layers.
    }
    \label{fig:grounding_acc}
    }
\end{floatrow}
\end{figure}

We conduct our ablation study on \nlvr and include two ablation models in \S \ref{experiment} and four additional variants of \model for comparison. 
For ease of computations, all these models are trained with only 36 features per image (including the full model).
Our analysis (Table \ref{nlvr_ablation}) aims to investigate the contributions of the following four components in \model:

\textbf{C1: Task-agnostic Pre-training.} 
We investigate the contribution of task-agnostic pre-training by entirely skipping such pre-training (\modelnp) and also by pre-training with only text but no images from COCO (VisualBERT w/o Grounded Pre-training).
Both variants underperform, showing that pre-training on paired vision and language data is important.

\textbf{C2: Early Fusion.} We include \modelby introduced in \S \ref{experiment} to verify the importance of allowing early interaction between image and text features, confirming again that multiple interaction layers between vision and language are important. 

\textbf{C3: BERT Initialization.}
All the models discussed so far are initialized with parameters from a pre-trained BERT model. 
To understand the contributions of the BERT initialization, we introduce a variant with randomly initialized parameters. 
The model is then trained as the full model.
While it does seem weights from language-only pre-trained BERT are important, performance does not degrade as much as we expect, arguing that the model is likely learning many of the same useful aspects about grounded language during COCO pre-training.

\textbf{C4: The sentence-image prediction objective.}
We introduce a model without the sentence-image prediction objective during task-agnostic pre-training (\model w/o Objective 2). Results suggest that this objective has positive but less significant effect, compared to other components.

Overall, the results confirm that the most important design choices are task-agnostic pre-training (C1) and early fusion of vision and language (C2).  In pre-training, both the inclusion of additional COCO data and using both images and captions are paramount.

\subsection{Dissecting Attention Weights}
\label{sec:analysis}

\label{sec:entitygrounding}

In this section we investigate which bounding regions are attended to by words, before \model is fine-tuned on any task.

\paragraph{Entity Grounding} First, we attempt to find attention heads within \model that could perform entity grounding, i.e., attending to the corresponding bounding regions from entities in the sentence. Specifically, we use the ground truth alignments from the evaluation set of Flickr30K. For each entity in the sentence and for each attention head in \model, we look at the bounding region which receives the most attention weight.
Because a word is likely to attend to not only the image regions but also words in the text, for this evaluation, we mask out the head's attention to words and keep only attention to the image regions.
Then we compute the how often the attention of a particular head agrees with the annotations in Flickr30K.

We report this accuracy\footnote{Despite that some heads are accurate at entity grounding, they are not actively attending to the image regions. For example, a head might be only allocating 10\% of its attention weights to all image regions, but it assigns the most of the 10\% weights to the correct region. We represent heads paying on average less than 20\% of its attention weights from the entities to the regions with smaller and light-colored dots and others with larger and bright dots.}, for all 144 attention heads in \model, organized by layer, in Figure~\ref{fig:grounding_acc}. We also consider a baseline that always chooses the region with the highest detection confidence. We find that \model achieves a remarkably high accuracy though it is not exposed to any direct supervision for entity grounding. The grounding accuracy also seems to improve in higher layers, showing the model is less certain when synthesizing the two inputs in lower layers, but then becomes increasingly aware of how they should align. We show examples of this behavior in \S \ref{sec:casestudy}.

\begin{figure*} [t]
\centering
\includegraphics[width=\textwidth]{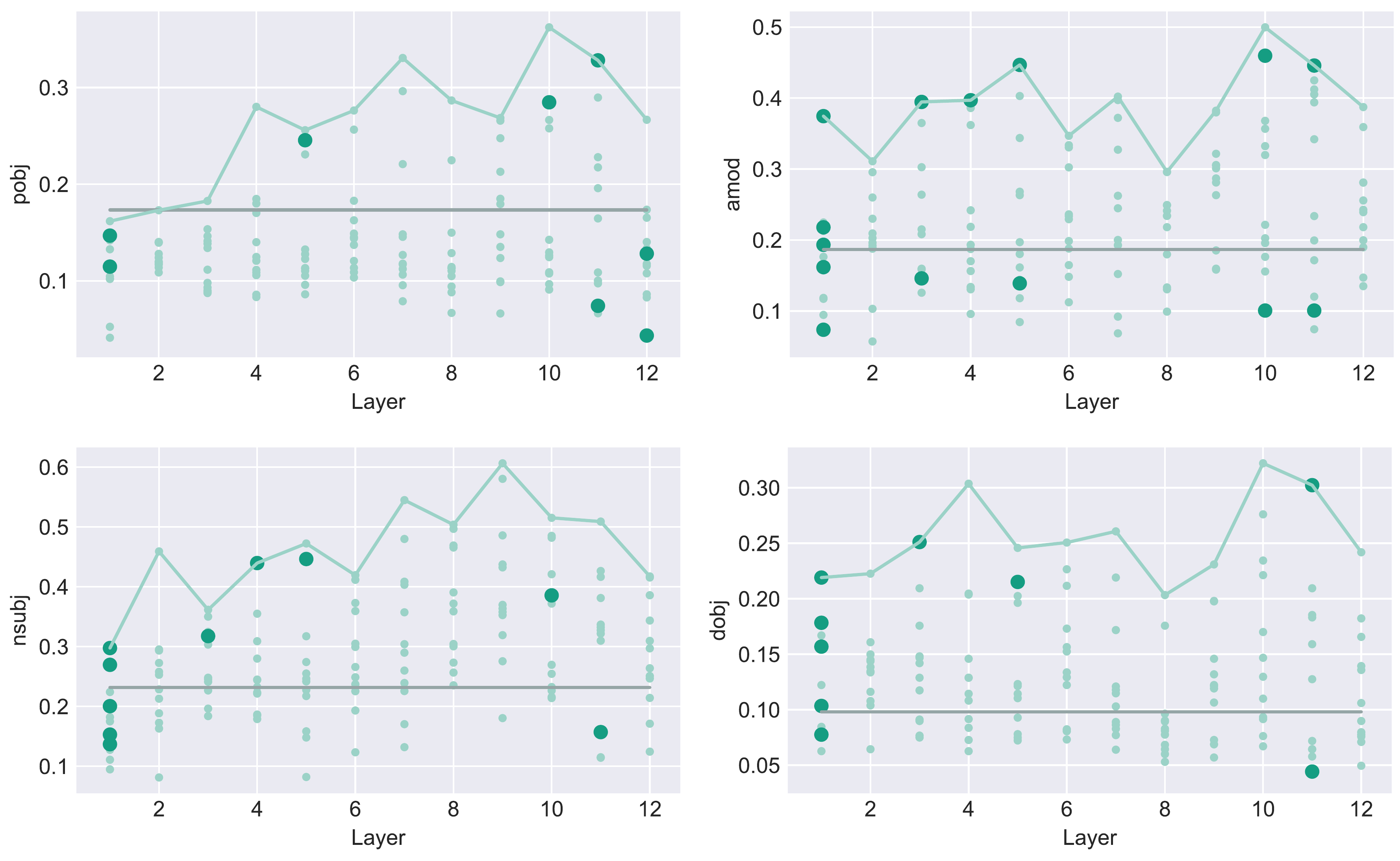}
\caption{Accuracy of attention heads of \model for predicting four specific dependency relationships (``pobj'', ``amod'', ``nsubj'', and ``dobj'') across modality. The grey lines denote a baseline that always chooses the region with the highest detection confidence. We observe that \model is capable of detecting these dependency relationships without direct supervision.}
\label{fig:syntactic}
\end{figure*}

\paragraph{Syntactic Grounding}
Given that many have observed that the attention heads of BERT can discover syntactic relationships~\citep{voita2019analyzing, clark2019does}, we also analyze how grounding information is passed through syntactic relationships that \model may have discovered. In particular, given two words that are connected with a dependency relation, $w_1 \stackrel{r}{\longrightarrow} w_2$, we would like to know how often the attention heads at $w_2$ attend to the regions corresponding to $w_1$, and vice-versa. For example, in Figure~\ref{fig:leading_example}, we would like to know if there is an attention head that, at the word ``walking'', is systematically attending to the region corresponding to the ``man'', because ``man'' and ``walking'' are related through a ``nsubj'' relation, under the Stanford Dependency Parsing formalism~\citep{de2008stanford}.

To evaluate such syntactic sensitivity in \model, we first parse all sentences in Flickr30K using AllenNLP's dependency parser~\citep{dozat2017deep,gardner2018allennlp}.
Then, for each attention head in \model, given that two words have a particular dependency relationship, and one of them has a ground-truth grounding in Flickr30K, we compute how accurately the head attention weights predict the ground-truth grounding. Examination of all dependency relationships shows that in \model, there exists at least one head for each relationship that significantly outperforms guessing the most confident bounding region. We highlight a few particularly interesting dependency relationships in Figure~\ref{fig:syntactic}
. Many heads seem to accurately associate arguments with verbs (i.e. ``pobj'', ``nsub'', and ``dobj'' dependency relations), arguing that \model is resolving these arguments, implicitly and without supervision, to visual elements. 

\subsection{Qualitative Analysis}
\label{sec:casestudy}
\begin{figure} [h]
\centering
\vspace{-20pt}
\includegraphics[width=0.95\textwidth]{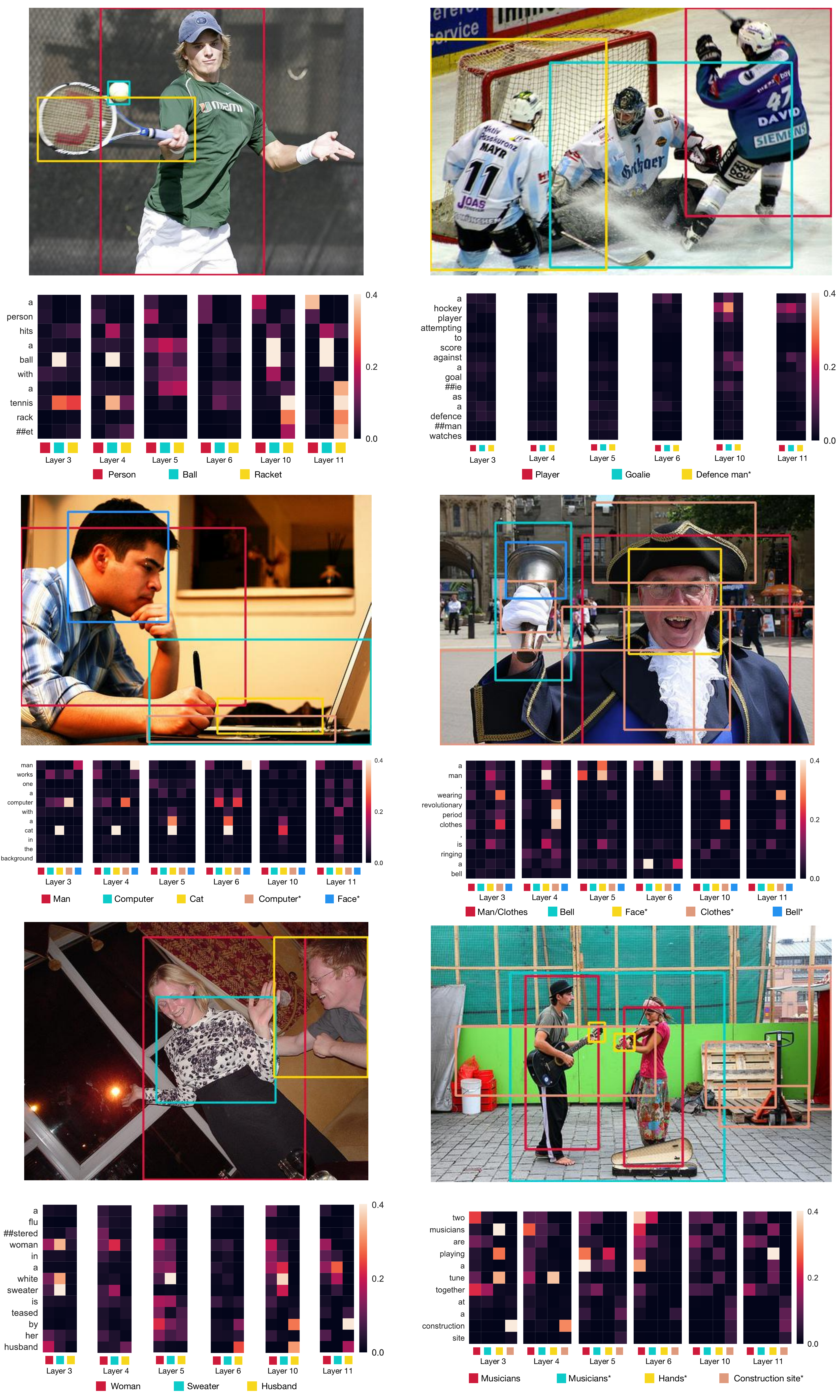}
\caption{Attention weights of some selected heads in \model on 6 examples. The first column is 3 random examples where alignments match Flickr30k annotations while the second column is 3 random examples where alignments do not match. }
\label{fig:examples}
\end{figure}

Finally, we showcase several interesting examples of how
\model changes its attention over the layers when processing images and text, in Figure \ref{fig:leading_example} and Figure~\ref{fig:examples}. To generate these examples, for each ground-truth box, we show a predicted bounding region closest to it and manually group the bounding regions into different categories.
We also include regions that the model is actively attending to, even if they are not present in the ground-truth annotation (marked with an asterisk).
We then aggregate the attention weights from words to those regions in the same category. We show the best heads of 6 layers that achieve the highest entity grounding accuracy.

Overall, we observe that \model seems to refine alignments through successive Transformer layers.
For example, in the bottom left image in Figure~\ref{fig:examples}, initially the word ``husband'' and the word ``woman'' both have significant attention weight on regions corresponding to the woman. 
By the end of the computation, \model has disentangled the woman and man, correctly aligning both.
Furthermore, there are many examples of syntactic alignments.
For example, in the same image, the word ``teased''  aligns to both the man and woman while ``by'' aligns to the man.
Finally, some coreference seems to be resolved, as, in the same image, the word ``her'' is resolved to the woman.

\section{Conclusion and Future Work}
In this paper, we presented \model, a pre-trained model for joint vision and language representation. Despite \model is simple, it achieves strong performance on four evaluation tasks. Further analysis suggests that the model uses the attention mechanism to capture information in an interpretable way. For future work, we are curious about whether we could extend \model to image-only tasks, such as scene graph parsing and situation recognition.  Pre-training \model on larger caption datasets such as Visual Genome and Conceptual Caption is also a valid direction.

\section*{Acknowledgement}
We would like to thank Xianda Zhou for help with experiments as well as Patrick H. Chen and members of UCLA NLP for helpful comments. We also thank Rowan Zellers for evaluation on VCR and Alane Suhr for evaluation on \nlvr.

\clearpage

\bibliography{iclr2019_conference}
\bibliographystyle{iclr2019_conference}

\newpage
\appendix
\section*{Implementation and Training details}
Below we introduce the implementation and training details for each evaluation task.

\section{VQA}\label{appendix:vqa}
Though the answers of VQA are open-ended, we follow the processing procedure of Pythia and consider it a classification problem, where the model only needs to choose one answer from a limited answer pool.
To better fit the pre-training process, a [MASK] token is appended after the question and the representation of the [MASK] token is fed into an output layer for classification.
Since there could be multiple correct answers to one question, we assign each correct answer with the same probability and minimize the cross entropy between the target probability and the output probability from the model,different from the binary cross entropy loss used in Pythia.
We do not use grid-level features from ResNet152 because it results in longer sequences and longer training time. 
\model (with/without Early Fusion) is pre-trained on COCO for 10 epochs with a batch size of 48 and a max learning rate of 5e-5.
For task-specific pre-training, all variants of \model are trained for 10 epochs with a batch size of 64 and a max learning rate of 5e-5.
Each input sequence consists of the question, the correct answer, and the image.
Only the masked language modeling with the image objective is used. 
During fine-tuning, they are trained with the task-specific objective for 10 epochs with a batch size of 64 and a max learning rate of 2e-5.
Following the practice in Pythia, for task-specific pre-training and fine-tuning, we train on the training and validation splits.

\section{VCR}\label{appendix:vcr}
In VCR, the task is decomposed into two subtasks, Q $\rightarrow$ A and QA $\rightarrow$ R.
For each sub-task, each training example contains four choices and we construct four input sequences, each containing the concatenation of the given question, a choice, and an image. When the model performs QA $\rightarrow$ R, the ``question'' part contains the original question and the correct choice, and the ``choice'' is a possible rationale.
The model is trained to classify which of the four input sequences is correct.

For \model (with/without Early Fusion), task-agnostic pre-training is conducted on COCO for 10 epochs with a batch size of 128 and a max learning rate of 1e-4. For all variants of \model, since R2C also performs task-specific pre-training with \bert on the VCR dataset for its text representation, we conduct task-specific pre-training with the same hyper-parameters (learning rate, batch size, optimizer warm-up ratio).
But notice that R2C conducts task-specific pre-training with text-only objective while we do so with a visually-grounded objective. 
During this step, each training example consists of a question, a choice, and an image.
Following R2C, we add an auxiliary task of predicting if the choice is correct. In the fine-tuning stage, for Q $\rightarrow$ A, we train for 8 epochs with a batch size of 32 and a max learning rate of 2e-5. For QA $\rightarrow$ R, we train for 12 epochs. For fine-tuning, we monitor the loss on the development set for early stopping.

\section{\nlvr}\label{appendix:nlvr}
For each training example in \nlvr, we construct a sequence consisting of the caption and image features from two images.
\model (with/without Early Fusion) is pre-trained on COCO for 10 epochs with a batch size of 64 and a max learning rate of 5e-5. 
For task-specific pre-training, similar to VCR, an auxiliary task is added to decide whether the caption in an training example is true.
All variants of \model are trained with the visually-grounded objective for a maximum of 10 epochs with a batch size of 64 and a max learning rate of 5e-5. In the fine-tuning stage, they are both trained for a maximum of 10 epochs with a batch size of 64 and a max learning rate of 5e-5.
For task-specific pre-training and fine-tuning, we monitor the loss on the development set for early stopping.

\section{Flickr30K}\label{appendix:flickr}
Since multiple boxes could be aligned to the same phrase, we use the same cross entropy loss used in our experiment on VQA, different from the binary cross entropy loss used in BAN.
\model (with/without Early Fusion) is pre-trained on COCO with a batch size of 32 and a learning rate of 5e-5.
During task-specific pre-training, all variants of \model are trained for 10 epochs with a batch size of 32 and a learning rate of 5e-5.
Only the masked language modeling with the image objective is used.
They are then fine-tuned with a maximum of 5 epochs with a batch size of 32 and a learning rate of 2e-5.
For task-specific pre-training and fine-tuning, we monitor the loss on the development set for early stopping.

\end{document}